\documentclass[journal]{IEEEtran}

\usepackage{amsmath,booktabs,graphicx,subfigure,float,amssymb}
\usepackage[sort&compress, numbers]{natbib}
\usepackage{hyperref}
\usepackage{tikz}
\usepackage{pgfplots}
\usepackage{pgfplotstable}

\usetikzlibrary{decorations.pathreplacing}
\usetikzlibrary{positioning, calc, arrows}

\def\X{{\bf X}}
\def\x{{\bf x}}
\def\Y{{\bf Y}}

\def\W{{\bf H}}
\def\G{{\bf G}}
\def\U{{\bf U}}
\def\V{{\bf V}}
\def\D{{\bf D}}
\def\G{{\bf G}}

\def\y{{\bf y}}

\begin{document}

\title{Multispectral and Hyperspectral Image Fusion Using a 3-D-Convolutional Neural Network}

\author{Frosti~Palsson,~\IEEEmembership{Student Member,~IEEE,}
        Johannes~R.~Sveinsson,~\IEEEmembership{Senior Member,~IEEE,}
				and Magnus~O.~Ulfarsson,~\IEEEmembership{Member,~IEEE}
        
\thanks{The authors are with the Faculty of Electrical and Computer Engineering, University of Iceland, Reykjavik,
E-mail: (frostip@gmail.com, sveinsso@hi.is, mou@hi.is)}
\thanks{This work was supported in part by the Icelandic Research Fund under Grant 174075-05 and in part by the University of Iceland Research Fund.}}


\maketitle

\begin{abstract}
In this paper, we propose a method using a three dimensional convolutional neural network (3-D-CNN) to fuse together multispectral (MS) and hyperspectral (HS) images to obtain a high resolution hyperspectral image. Dimensionality reduction of the hyperspectral image is performed prior to fusion in order to significantly reduce the computational time and make the method more robust to noise. Experiments are performed on a data set simulated using a real hyperspectral image. The results obtained show that the proposed approach is very promising when compared to conventional methods. This is especially true when the hyperspectral image is corrupted by additive noise.
\end{abstract}


\section{Introduction}
\label{Sec:intro}

Pansharpening, which is the fusion of a multispectral (MS) image and a wide-band panchromatic (PAN) image, is an important technique in remote sensing. Ideally, the fused image should contain all the spectral information from the MS image and all the spatial details from the PAN image.
 
With advances in sensor development, the fusion of a high spatial resolution MS image and a low spatial resolution hyperspectral (HS) image (MS/HS fusion) is becoming relevant for many applications. A typical HS image contains hundreds of spectral reflectance bands, making the spectral information content very high. This allows for the identification of different materials based on their spectral signature, which is useful for applications such as classification of land cover types.

Numerous pansharpening methods have been proposed in recent years and they are often categorized either as component substitution (CS) or multi-resolution analysis (MRA) methods. The CS and MRA methods can generally be described using a simple detail injection framework \cite{critical}. Apart from the CS and MRA methods there are model based methods such as \cite{TV,sparsepriors}, and methods based on statistical inference such as \cite{Yifan,pwmbf}.

Although MS/HS fusion is a relatively new topic in remote sensing, there are several publications on the topic, including \cite{MSHS_1}, which used sparse coding and spectral unmixing. A method using coupled non-negative matrix factorization and spectral unmixing was given in \cite{MSHS_2}. In \cite{MSHS13}, a method based on a 3D-wavelet transform was proposed. A common approach to the MS/HS fusion problem is to view it as a number of pansharpening sub-problems where each spectral band of the MS image takes the role of the PAN image \cite{MSHS11,MSHS7}.

In the past decade, methods based on Deep Learning (DL) have in many cases outperformed traditional signal processing approaches in areas such as speech and pattern recognition \cite{szegedy2015going}. The main component of DL is the artificial neural network (NN). More specifically, the so-called convolutional neural network (CNN) has been shown to be effective in pattern recognition and related areas \cite{le1990handwritten}. 

DL based methods have previously been used to solve the pansharpening problem \cite{DLfusion1,DLfusion2}. Here, we propose a MS/HS fusion method using DL.

The method is based on training a 3D-CNN for learning filters used to fuse the MS and HS images. Since the method is based on supervised learning, it requires a target HS image, which is not available. Therefore, the input data need to be spatially decimated (low-pass filtered and downsampled) in order to use the observed HS image as the target image. The assumption being made here is that the relationship between the input and target data, learned by the 3D-CNN at a lower resolution scale, also applies for a higher resolution scale.

To make the fusion problem more computationally efficient, the dimensionality of the HS image is reduced prior to the fusion stage using principal component analysis (PCA) \cite{PCAjolliffe}. This is an important step of the proposed method and it depends on the assumption that the spectral singular vectors of the lower resolution HS image are identical to those of the higher resolution HS image that we want to estimate. By comparing our approach to the conventional methods given in \cite{Yifan,pwmbf}, it is demonstrated that the proposed method gives better results according to three quantitative quality metrics. 

Another advantage of the proposed method is that the 3D-CNN learns the decimation filter in an automatic manner. In other words, the method is relatively insensitive to the choice of decimation filter used to prepare the training samples for the 3D-CNN. It also produces images that are free of artifacts, such as halos and ringing artifacts, often seen when using conventional methods.

The outline of this paper is as follows. In the next section, we briefly discuss CNNs. In Section III, the proposed method is described in detail. In Section IV, we present experimental results, and finally, in Section V, the conclusion is drawn. 

\section{Convolutional Neural Networks}
\label{Sec:CNN}

CNNs consist of convolutional layers and a convolutional layer consists of a number of hidden layers that contain a number of neurons. The main idea behind CNNs is the concept of a local receptive field \cite{hubel1962receptive}, that is associated with each neuron in a hidden layer. The input to a convolutional layer is an image of one or more channels. Each neuron in a hidden layer receives input from a rectangular subset of the input image, which is the neuron's receptive field.

By sliding the receptive field over the input image, and after each shift connecting to a new neuron in the hidden layer, the neurons of the hidden layer provide a complete tiling of the input image. All the neurons in the hidden layer share their weights and bias and therefore they can detect the same feature at different locations in the input image.

The output of a hidden layer is called a feature map and the shared weights are called a filter. A single convolutional layer can have many such feature maps and hence can learn several different filters that detect different distinct features. Between convolutional layers there can be so-called pooling layers, which sub-sample the feature maps according to some function, e.g.,  maximum value (max-pooling). This simplifies the feature maps and greatly reduces the number of parameters in the network as the number of convolutional layers grows.

The main benefit of the CNN architecture is that much fewer parameters (weights and biases) need to be learned than for a conventional fully connected NN. This is due to the shift-invariance, i.e., the shared weights of the locally connected neurons in the hidden layers of the CNN and enables the construction of deeper networks that can learn much faster, without sacrificing performance.

In a 3D-CNN, which has 3D filters and 3D receptive fields, the output of the $n$th feature map $\y^n$ at location $\{i,j,k\}$ is given by
$$y^n_{i,j,k} = \sigma(b^n+(\W^n*\x)_{i,j,k}),$$
where $*$ denotes 3D-convolution, $b^n$ and $\W^n$ are the shared bias and filter (shared weights), respectively, $\sigma$ denotes the non-linear activation function and $\x$ is the input.

\usetikzlibrary{patterns}
\usetikzlibrary{shapes.misc}
\begin{figure}
\begin{center}
\def\Dl{L}
\def\mh{1}
\def\mw{1.3}
\begin{tikzpicture}[ultra thick, scale = 0.68, transform shape,font=\normalsize]
\tikzstyle{myarrows}=[line width=0.3 mm,draw=blue,-triangle 45,postaction={draw, line width=0.2mm, shorten >=2mm, -}]
\node[rectangle] (a) [draw, dashed, thick,minimum width=6.5cm,minimum height=3cm,xshift=-1.8cm,yshift=-0.8cm] {};
\node[rectangle] (b) [draw, dashed, thick,minimum width=6.5cm,minimum height=2cm,yshift=-0.5cm,xshift=0.0cm,below of=a,node distance=3cm] {};
\node[rectangle] (a1) [draw, thick,minimum width=1.5cm,minimum height=1.5cm,anchor=north,pattern=grid,pattern color=gray!45] {$\G^r$} ;
\coordinate[above=0.3cm of a1]  (t1)  {};
\node at (t1) {$\X^\text{HS}=\G\U^T$};
\node[rectangle] (a2) [draw, thick, minimum width=2.5cm,minimum height=2.5cm,left of=a1,node distance=3cm] {$\X^\text{MS}$};
\node[rectangle] (a3) [draw, thick,minimum width=1.5cm,minimum height=1.5cm,below of=a1,node distance=3.5cm] {$\tilde{\G}_\text{LR}^r$};
\node[rectangle] (a4) [draw, thick,minimum width=1.5cm,minimum height=1.5cm,below of=a2,node distance=3.5cm] {$\X^\text{MS}_\text{LR}$};
\node[rectangle] (a5) [draw, thick,minimum width=1.5cm,minimum height=1.5cm,below of=a4,node distance=3cm] {}; 
\node[rectangle] (a6) [draw,fill=white,pattern=grid,pattern color=gray!45, thick,minimum width=1.5cm,minimum height=1.5cm,below of=a4,node distance=3cm,xshift=0.2cm,yshift=-0.2cm] {$\X_\text{LR}$};
\node[rounded rectangle] (CNN) [draw, thick, minimum width=2.5cm,minimum height=1.5cm,right of=a6,node distance=3cm,text width=1.5cm,text centered] {supervised training} ;
\coordinate[left=0.5cm of a]  (i1)  {};
\coordinate[left=0.5cm of b]  (i2)  {};
\coordinate[left=1cm of a6]  (i3)  {};
\node at (i1) {\Large 1)};
\node at (i2) {\Large 2)};
\node at (i3) {\Large 3)};
\coordinate[right=0.05cm of a1]  (c1)  {};
\coordinate[right=1.2cm of c1]   (c2)  {};
\coordinate[below=5.05cm of c2]   (c3)  {};
\coordinate[above=0.9cm of CNN]   (c3a)  {};
\coordinate[above=0.05cm of CNN]   (c4)  {};
\coordinate[right=0.05cm of a6]  (c5)  {};
\coordinate[left=0.05cm of CNN]  (c6)  {};
\coordinate[below=0.2cm of a4]  (c7)  {};
\coordinate[above=0.05cm of a5]  (c8)  {};
\coordinate[below=0.05cm of a]  (c9)  {};
\coordinate[above=0.05cm of b]  (c10)  {};
\node[rectangle] (tCNN) [draw, line width=0.5mm,minimum width=1.5cm,minimum height=1.5cm,right of=CNN,node distance=3.5cm,text width=1.5cm,text centered] {trained 3D-CNN};
\coordinate[right=0.05cm of CNN]  (c11)  {};
\coordinate[left=0.05cm of tCNN]  (c12)  {};
\coordinate[below=0.3cm of a6]  (t2)  {};
\coordinate[below=0.8cm of a6]  (t3)  {};
\draw [myarrows](c1)--(c2)--(c3)--(c3a)--(c4) node[above=0.1cm,midway] {};
\draw [draw=none](c2)--(c3) node[right=0.3cm,midway,rotate=-90,xshift=-1cm] {target patches};
\draw [myarrows](c5)--(c6) node[above=0.1cm,midway] {};
\draw [myarrows](c7)--(c8) node[above=0.1cm,midway] {};
\draw [myarrows](c9)--(c10) node[above=0.1cm,midway] {};
\draw [myarrows](c11)--(c12) node[above=0.1cm,midway] {};
\node at (t2) {input patches};
\node at (t3) {$\X_\text{LR}=[\X^\text{MS}_\text{LR}~\tilde{\G}_\text{LR}^r]$};
\end{tikzpicture}
\end{center}
\caption{General outline of the training part of the algorithm. The steps labeled 1), 2) and 3), correspond to similarly labeled steps in the text.}
\label{fig:outline}
\end{figure}
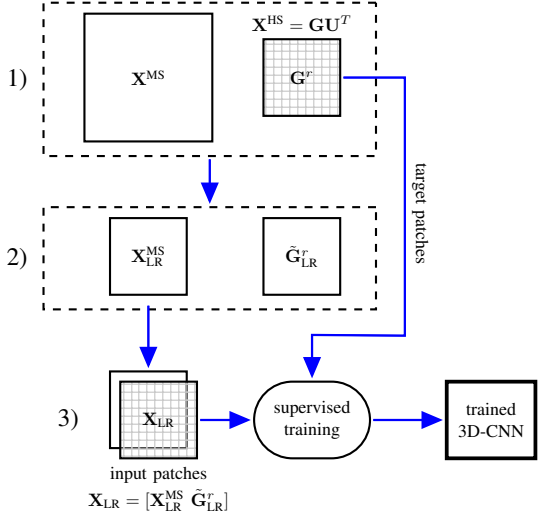

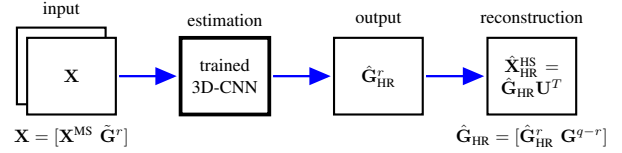
\begin{figure}
\begin{center}
\begin{tikzpicture}[ultra thick, scale = 0.68, transform shape,font=\normalsize]
\tikzstyle{myarrows}=[line width=0.3 mm,draw=blue,-triangle 45,postaction={draw, line width=0.2mm, shorten >=2mm, -}]
\node[rectangle] (i1) [draw, thick,minimum width=1.5cm,minimum height=1.5cm,node distance=3.5cm,text width=1.5cm,text centered] {};
\node[rectangle] (input) [draw, fill=white, thick,minimum width=1.5cm,minimum height=1.5cm,node distance=3.5cm,text width=1.5cm,text centered,xshift=0.2cm,yshift=-0.2cm] {$\X$};
\node[rectangle] (tCNN) [draw, line width=0.5mm,minimum width=1.5cm,minimum height=1.5cm,right of=input,node distance=3cm,text width=1.5cm,text centered] {trained 3D-CNN};
\coordinate[right=0.05cm of input]  (c1)  {};
\coordinate[left=0.05cm of tCNN]   (c2)  {};
\draw [myarrows] (c1)--(c2) node[above=0.1cm,midway] {};
\node[rectangle] (output) [draw, thick,minimum width=1.5cm,minimum height=1.5cm,right of=tCNN,node distance=3cm,text width=1.5cm,text centered] {$\hat{\G}_{\text{HR}}^r$};
\coordinate[right=0.05cm of tCNN]  (c3)  {};
\coordinate[left=0.05cm of output]   (c4)  {};
\draw [myarrows] (c3)--(c4) node[above=0.1cm,midway] {};
\node[rectangle] (hr) [draw, thick,minimum width=1.5cm,minimum height=1.5cm,right of=output,node distance=3cm,text width=1.5cm,text centered] {$\hat{\X}_\text{HR}^\text{HS}=\hat{\G}_\text{HR}\U^T$};
\coordinate[below=0.4cm of input]  (t1)  {};
\node at (t1) {$\X=[\X^{\text{MS}}~\tilde{\G}^r]$};
\coordinate[below=0.4cm of hr]  (t2)  {};
\node at (t2) {$\hat{\G}_\text{HR}=[\hat{\G}_{\text{HR}}^r~\G^{q-r}]$};
\coordinate[above=0.4cm of hr]  (t3)  {};
\node at (t3) {reconstruction};
\coordinate[right=0.05cm of output]  (c5)  {};
\coordinate[left=0.05cm of hr]   (c6)  {};
\draw [myarrows] (c5)--(c6) node[above=0.1cm,midway] {};
\node[rectangle] (t4) [draw=none,above of=i1,node distance=1.1cm] {input};
\node[rectangle] (t5) [above of=output,node distance=1.1cm] {output};
\node[rectangle] (t6) [above of=tCNN,node distance=1.1cm] {estimation};
\end{tikzpicture}
\end{center}
\caption{General outline of estimation part of the algorithm. The trained CNN is fed the entire input data at its full resolution and yields the high resolution spatial loadings, which are used to reconstruct the estimated high resolution HS image via the inverse PCA transform.}
\label{fig:estimation}
\end{figure}

\section{Proposed Method}
\label{Sec:method}

In this section, we first describe the proposed method and then discuss the chosen architecture of the 3D-CNN. 
\subsection{General Outline of the Method}
\label{SSec:outline}
The observed MS image is denoted by $\X^{\text{MS}}$ and is of dimension $M \times N\times P$, where $P$ is the number of spectral bands. The observed $m \times n\times q$ HS image is denoted by $\X^{\text{HS}}$. The training of the 3D-CNN is performed as shown in Fig. \ref{fig:outline} and described below. 

To simplify the notation, we use the same symbols for 3D-matrices, i.e., images, and normal matrices. The implicit reshaping of a 3D image into a matrix with vectorized images (bands) in the columns is assumed. A tilde above a symbol denotes interpolation (upsampling followed by spatial filtering), a hat above a symbol denotes an estimate, and concatenation/stacking of matrices/images is denoted by square brackets, e.g., $[\X ~\Y]$. 

\begin{enumerate}
	\item  Dimensionality reduction of $\X^\text{HS}$ using PCA. Singular value decomposition gives
$$\X^\text{HS}=\V\D\U^T=\G\U^T,$$
where the $mn \times q$ matrix $\G=\V\D$ contains the spatial loadings and the $q\times q$ matrix $\U$ contains the spectral singular vectors. The first $r$ columns of $\G$ are used to form an $m\times n\times r$ image, $\G^r$.
  
	\item The image $\X^{\text{MS}}$ is spatially decimated by the resolution factor between the MS and HS images, using a bicubic decimation filter, to yield $\X_{\text{LR}}^{\text{MS}}$, of dimension $m \times n\times P$. Similarly, $\G^r$, is spatially decimated and then interpolated (using a bicubic filter) to the dimension of $\X_{\text{LR}}^{\text{MS}}$, yielding $\tilde{\G}_{\text{LR}}^r$.
	
	\item $\X_\text{LR}^\text{MS}$ and $\tilde{\G}_\text{LR}^r$ are stacked to obtain the $m\times n\times (r+P)$ input image $\X_{\text{LR}}=[\X_\text{LR}^\text{MS}~\tilde{\G}_\text{LR}^r]$. The target data are the first $r$ bands of $\G$, i.e., $\G^r$. $\X_{\text{LR}}$ and $\G^r$, are randomly divided into a number of matching patches (overlapping) of size ${7\times 7}$ pixels, of depth $r+P$ and $r$, for inputs and targets, respectively.
\end{enumerate}

The fusion part of the method is depicted in Fig. \ref{fig:estimation}. The trained 3D-CNN can accept the entire input data at once, without having to break it down into patches, since it has learned all its filters. The input $\X$ to the trained 3D-CNN consists of the stacked MS image, $\X^\text{MS}$, and the first $r$ spatial loadings of $\G$, which have been interpolated to the size of $\X^\text{MS}$, i.e., $\X=[\X^\text{MS}~\tilde{\G}^r]$. The output of the 3D-CNN is the estimated high resolution loadings, $\hat{\G}_\text{HR}^r$. 

The final step is the reconstruction of the estimated high resolution HS image, $\hat{\X}_{\text{HR}}^\text{HS}$, via 
$$\hat{\X}_{\text{HR}}^\text{HS}=\hat{\G}_\text{HR}\U^T=[\hat{\G}_\text{HR}^r~\tilde{\G}^{q-r}]\U^T,$$ 
where $\hat{\G}_\text{HR}=[\hat{\G}_\text{HR}^r~\tilde{\G}^{q-r}]$ and $\tilde{\G}^{q-r}$ are the remaining $q-r$ interpolated spatial loadings obtained from the observed HS image, $\X^\text{HS}$, and the matrix $\U$ contains the spectral singular vectors of $\X^\text{HS}$. 

There are two options for the reconstruction of the estimated fused image. The first option is the one described above, where the first $r$ loadings in $\tilde{\G}$ are replaced by the high resolution estimate $\hat{\G}_{\text{HR}}^r$. If the HS image is noisy, a second option is to retain only the first $r$ PCs, i.e., performing the inverse PCA transform using the reduced $\hat{\G}_{\text{HR}}^r$ and $\U^r$ matrices, yielding $\hat{\X}_{\text{HR}}^\text{HS}=\hat{\G}_{\text{HR}}^r{\U^r}^T$, where $\U^r$ denotes the reduced matrix $\U$. 

\subsection{CNN Architecture}
\label{SSec:CNN}

In this work, we have decided to use a 3D-CNN architecture since an HS image has two spatial dimensions and one spectral dimension, and a 3D-CNN learns spectral-spatial features. If the input to a convolutional layer is an $M\times N\times P$ image, and the filter size is $i\times j\times k$, the resulting feature map is of size $M-i+1\times N-j+1\times P-k+1$. To preserve the size of the input image through the layers of the 3D-CNN, and to avoid boundary artifacts due to the convolution operations, the input to a convolutional layer with filter size $i\times j\times k$, needs to be zero-padded by $(i-1)/2$ zeros at each end of the first dimension, $(j-1)/2$ for the second dimension, and $(k-1)/2$ for the third dimension. 

The 3D-CNN used in our experiments has 3 convolutional layers with 32, 64 and $r$ filters, respectively, where $r$ is the number of spatial loadings to sharpen. The filter sizes for the first two convolutional layers were chosen equal to $3\times 3\times 3$, and $1\times 1\times 1$ for the last convolutional layer. Each convolutional layer is preceded by a zero-padding layer and followed by a Gaussian noise regularization layer (except for the output layer), which adds zero-mean Gaussian noise to the output of the previous layer. This helps to reduce overfitting in the network, and is a form of random data augmentation \cite{GNlayer}. The first two convolutional layers have rectified linear unit (ReLU) activation functions, i.e., $\sigma(x)=\text{max}(x,0)$, while the output layer has linear activation. 

The input shape for the convolutional layers is flexible. In other words, the 3D-CNN can be trained using a specific patch size, and when the CNN has been trained, the entire input can be estimated at once. This can be very memory consuming if the input image is large, and therefore PCA dimensionality reduction helps to significantly reduce the memory overhead in the fusion process. The layers of the CNN are summarized in Table \ref{tab:CNN}.

\begin{table}[htbp]
  \centering
  \caption{3D-CNN architecture. The numbers in parenthesis following zero-padding layers indicate the number of zeros added to each dimension. The numbers in parenthesis after convolution3D, indicate number of filters and the filter size of each dimension. $r$ indicates the number of PCs. Finally, the number following Gaussian noise denotes the noise variance.}
    \begin{tabular}{clc}
    \addlinespace
    \toprule
    layer \# & Type & Activation \\
    \midrule
    1 & zero-padding3D (1,1,1) & none \\
    2 & convolution3D (\textbf{32},3,3,3) & ReLU \\
    3 & Gaussian noise (0.5) & none \\
    4 & zero-padding3D (1,1,1) & none \\
    5 & convolution3D (\textbf{64},3,3,3) & ReLU \\
    6 & Gaussian noise (0.5) & none \\
    7 & convolution3D ($\bf r$,1,1,1) & none  \\
    \bottomrule
    \end{tabular}%
  \label{tab:CNN}%
\end{table}%

\begin{figure}[htbp]
	\centering
		\includegraphics[width=0.7\linewidth]{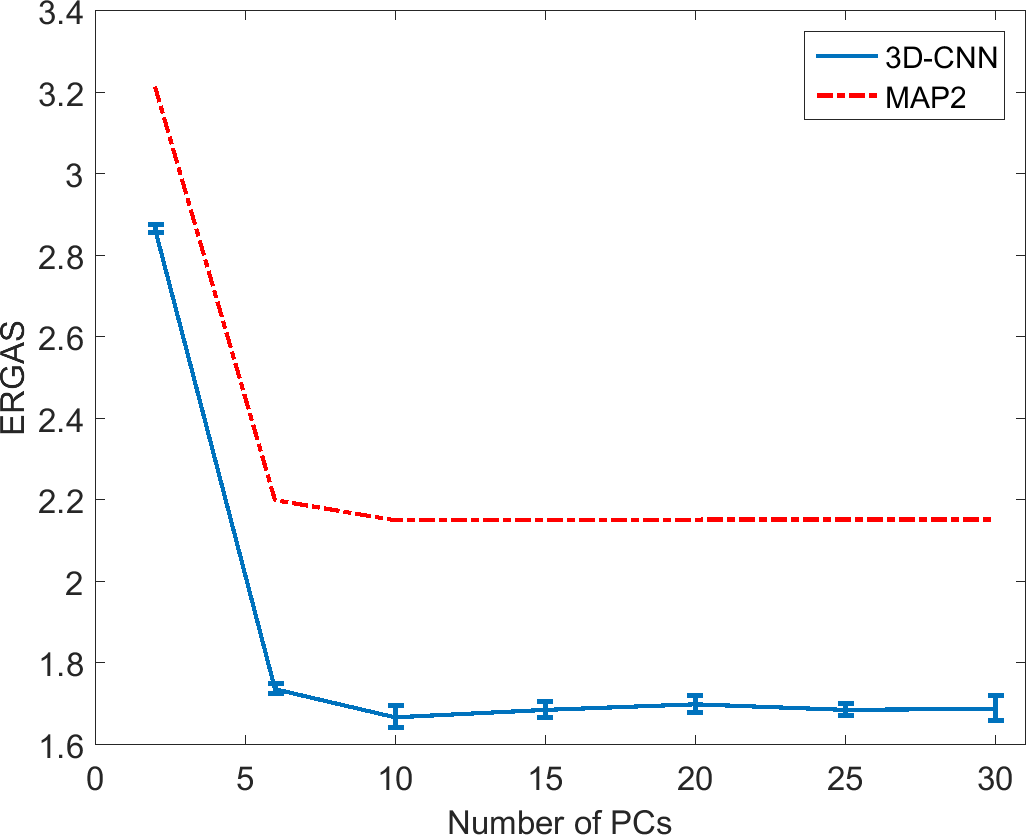}
	\caption{Performance in terms of ERGAS of the proposed and MAP2 methods, as a function of the number of PCs. Six trials were performed for the 3D-CNN method. The mean is shown and the standard deviation is displayed using errorbars.}
	\label{fig:PCA_ERGAS}
\end{figure}

\section{Experiments}
\subsection{Simulated Data}

The HS image used in the experiments is the ROSIS Pavia center dataset \footnote{The ROSIS data was kindly provided by Prof. P. Gamba from the University of Pavia, Italy.}. Its dimension is $512\times 512$ pixels with 102 spectral bands. However, there is a blank strip along the left side and thus we only use 480 pixels along the row dimension. 

The MS image is simulated from the HS image by averaging bands of the HS image according to the spectral response profiles of the R, G, B and NIR bands of the IKONOS MS sensor. We spatially decimate the observed HS image by a factor of 4 using a bicubic decimation filter to obtain the lower spatial resolution HS image. This yields the images to be fused, i.e., an MS image of dimension $512\times 480$ pixels with 4 spectral bands, and an HS image of dimension of $128\times 120$ pixels and 102 spectral bands. The original HS image is used as the reference image for the quantitative quality evaluation. 

The method was implemented in the Python programming language using the Keras DL library which runs on top of the Theano backend \footnote{\url{http://keras.io},~~\url{https://github.com/Theano}} and the computations were performed using an Intel i5-2400 CPU@3.1 GHz with 16GB of RAM. 

\subsection{Results}

As described in Section \ref{SSec:outline}, the simulated MS image and HS spatial loadings are spatially decimated and used as the input to the 3D-CNN. The target data are the first $r$ spatial loadings of the HS image. The training data consist of 8192 randomly chosen and matched patches of spatial size $7\times 7$ pixels, from the input and target data.

The network objective function is the mean squared error (MSE) between the target patch and the estimated patch. We use the adaptive moment estimation (ADAM) \cite{kingma2014adam} optimizer and we use the values for the optimizer parameters as given in \cite{kingma2014adam}. The number of training epochs is equal to 50, and it was assured that the objective function had fully converged during the training. Finally, the batch size is equal to 5, and the variance for the Gaussian noise regularization layer is equal to 0.5. The batch size is the number of samples that are propagated through the CNN at each time. We found that a small batch size gives better results and faster convergence.

The proposed method is compared to the method in \cite{Yifan}, and the extended version of that method in \cite{pwmbf}. We refer to these methods as MAP1 and MAP2, respectively. Both comparison methods are based on maximum a posteriori (MAP) estimation of wavelet coefficients, while the MAP2 method is identical to MAP1, except for PCA dimensionality reduction, in a similar manner to the proposed method.

We begin by investigating the effect of the number of sharpened PCs (spatial loadings) on the performance of the proposed and MAP2 methods in terms of the ERGAS \cite{ERGAS} metric. The following number of PCs are considered: 2, 6, 10, 15, 20, 25 and 30. The results of this experiment are shown in Fig. \ref{fig:PCA_ERGAS}. According to the figure, 10 PCs give optimal results for both methods. In the following experiments, 10 PCs are used for these methods.

The next experiment is the evaluation of the fusion performance for all methods in terms of the ERGAS \cite{ERGAS}, SAM \cite{SAM} and SSIM \cite{SSIM} quantitative quality metrics, without and with additive zero-mean Gaussian noise (SNR=20dB). The results of the quantitative quality evaluation are summarized in Table \ref{tab:res1}. The upper half of the table gives the results without added noise. As shown there, the proposed method significantly outperforms the MAP1 and MA2 methods according to all three quality metrics. Of the comparison methods, MAP2 performs much better than MAP1. Obviously, the proposed method is more costly than the comparison methods in terms of computation time. By using a powerful graphical processing unit (GPU), the training time could be reduced by up to the order of magnitude 2, making the proposed method competitive in terms of computation time. Fig. \ref{fig:band80} depicts only a small portion of the 102th band of the interpolated, reference, and estimated HS image for all methods. Visual inspection shows that the proposed method gives the best results.

The lower half of Table \ref{tab:res1} summarizes the results obtained with a zero-mean Gaussian noise added to the HS image. Again, the proposed method performs significantly better in terms of the quality metrics. However, its noise tolerance is similar, or slightly less than for the MAP2 method. The MAP1 method, which does not use PCA prior to the fusion, performs significantly worse than the other methods in the presence of noise. 

Next, the performance of all methods, in terms of the ERGAS metric, is compared when the SNR varies due to additive Gaussian noise, from 10 to 30 dB, in increments of 5 dB. The result of this experiment is shown in Fig. \ref{fig:SNR}. The plot clearly emphasizes what was observed in the previous experiment. The proposed method performs best and the MAP1 method performs significantly worse. 

Finally, the sensitivity of all methods w.r.t. the decimation filter used, is investigated. Three types of decimation filters are considered, i.e., bicubic, bilinear and nearest neighbor. The results for the methods measured by the ERGAS and SAM metrics are summarized in Table \ref{tab:filters}. According to the table, bicubic decimation gives the best results for all the methods. Using bilinear decimation degrades the performance of the methods in terms of the ERGAS and SAM metrics, however, the proposed method and MAP2 are less affected than the MAP1 method. Finally, nearest neighbor decimation degrades the performance significantly more for the MAP1 and MAP2 methods, than for the proposed method, when compared to the results obtained using the bicubic decimation.

\def\wi{0.76}
\begin{figure}[htpb]
\begin{center}
\includegraphics[width=\wi\linewidth]{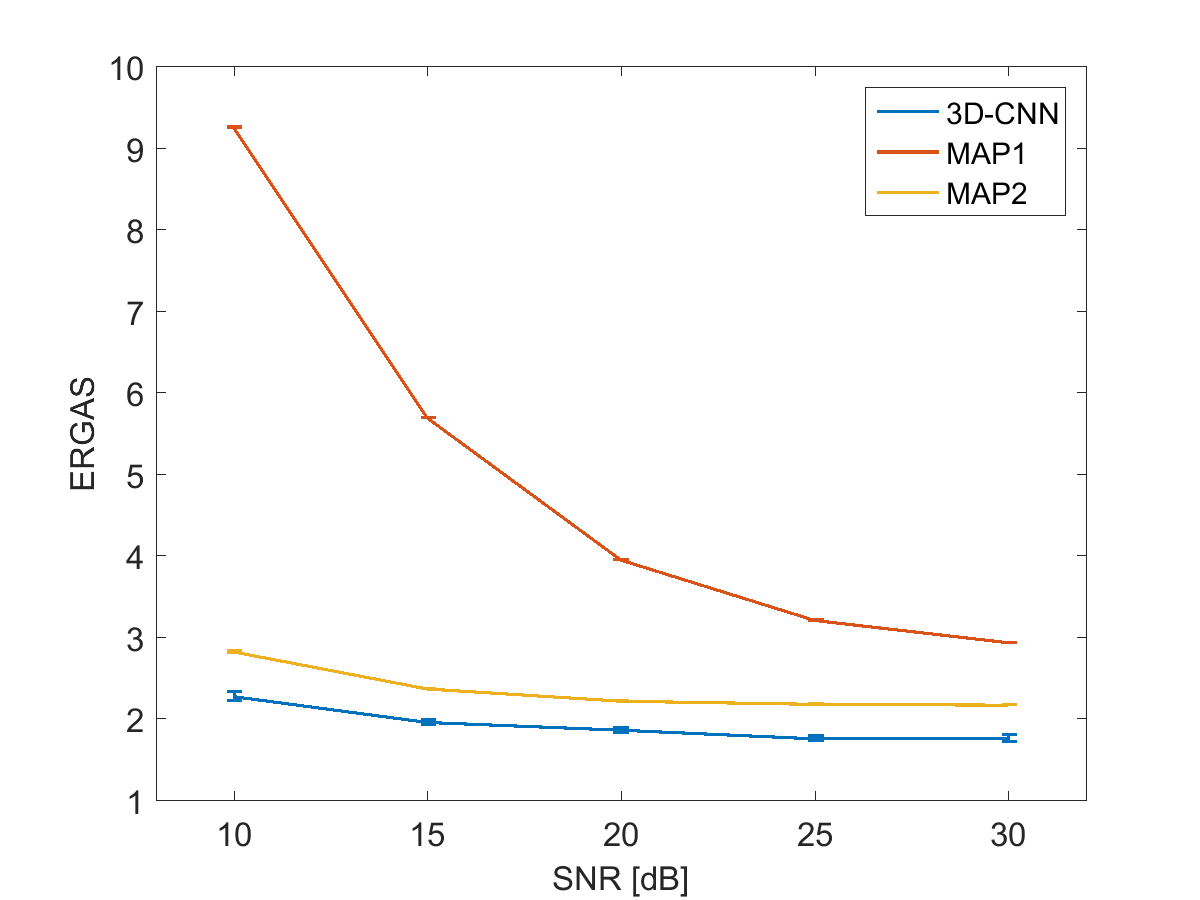}
\caption{Noise resistance of the proposed method vs comparison methods. For each value of SNR, 6 trials were conducted and the graph shows the mean and standard deviation of the trials as a function of the SNR.}
\label{fig:SNR}
\end{center}
\end{figure}

\def\cmrw{1pt}
\begin{table}[htbp]
  \centering
  \caption{Quantitative quality evaluation results, without and with additive Gaussian noise (SNR=20dB). For the proposed and MAP1 methods, 10 PCs were used in reduced PCA. The CPU time is given in seconds. Six trials were performed for the noisy case, and also for the proposed method without noise.}
    \begin{tabular}{cccccc}
    \addlinespace
    \toprule
    Method  & ERGAS & SAM & SSIM & CPU time\\
    \cmidrule[\cmrw](lr){1-1} \cmidrule[\cmrw](lr){2-2} \cmidrule[\cmrw](lr){3-3} \cmidrule[\cmrw](lr){4-4} \cmidrule[\cmrw](lr){5-5}
    MAP1\cite{Yifan}  &2.806&3.711&0.971& 45 \\
    MAP2\cite{pwmbf}  &2.17&3.26&0.978& 9\\
    3D-CNN  & \textbf{1.676$\pm$0.02}& \textbf{2.730$\pm$0.02} & \textbf{0.988$\pm$1.14e-4} &978$\pm$8 \\
		\midrule
    \multicolumn{5}{c}{Noisy HS image (SNR=20dB)} \\
		\midrule
    MAP1\cite{Yifan}  & 3.95$\pm$0.004  & 7.42$\pm$0.005 & 0.89$\pm$6.98e-4 & 45$\pm$0.4\\
    MAP2\cite{pwmbf} & 2.23$\pm$0.002 & 3.46$\pm$0.003 & 0.98$\pm$6.06e-5 & 9$\pm$0.4\\
    3D-CNN & \textbf{1.79$\pm$0.05} & \textbf{3.03$\pm$0.02} & \textbf{0.99$\pm$3.3e-4}& 989$\pm$10 \\
    \bottomrule
    \end{tabular}%
  \label{tab:res1}%
\end{table}%

\begin{table}[htbp]
  \centering
  \caption{Performance of all methods w.r.t. to the interpolation filter used. Bicubic, bilinear and nearest neighbor interpolation is considered. One trial was performed for the proposed method.}
    \begin{tabular}{ccccccc}
    \addlinespace
    \toprule
     & \multicolumn{2}{c}{Bicubic} & \multicolumn{2}{c}{Bilinear} & \multicolumn{2}{c}{Nearest} \\
     \cmidrule[\cmrw](lr){2-3} \cmidrule[\cmrw](lr){4-5} \cmidrule[\cmrw](lr){6-7}
    Method & ERGAS & SAM & ERGAS & SAM & ERGAS & SAM \\
		\midrule
    MAP1 & 2.806 & 3.711 & 3.080 & 4.721 & 5.680 & 5.501 \\
    MAP2 & 2.170 & 3.260 & 2.233 & 3.468 & 5.234 & 5.193 \\
    3D-CNN & 1.676 & 2.730 & 2.069 & 3.022 & 3.104 & 3.858 \\
    \bottomrule
    \end{tabular}%
  \label{tab:filters}%
\end{table}%

\def\wi{0.19}
\begin{figure*}[htpb]
\begin{center}
\subfigure[Interpolated HS image]{\includegraphics[width=\wi\linewidth]{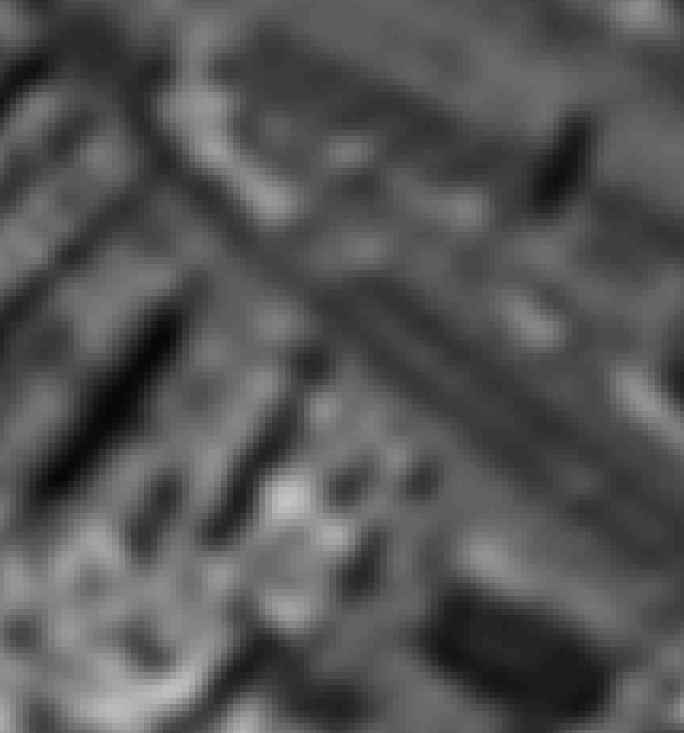}}
\subfigure[Reference]{\includegraphics[width=\wi\linewidth]{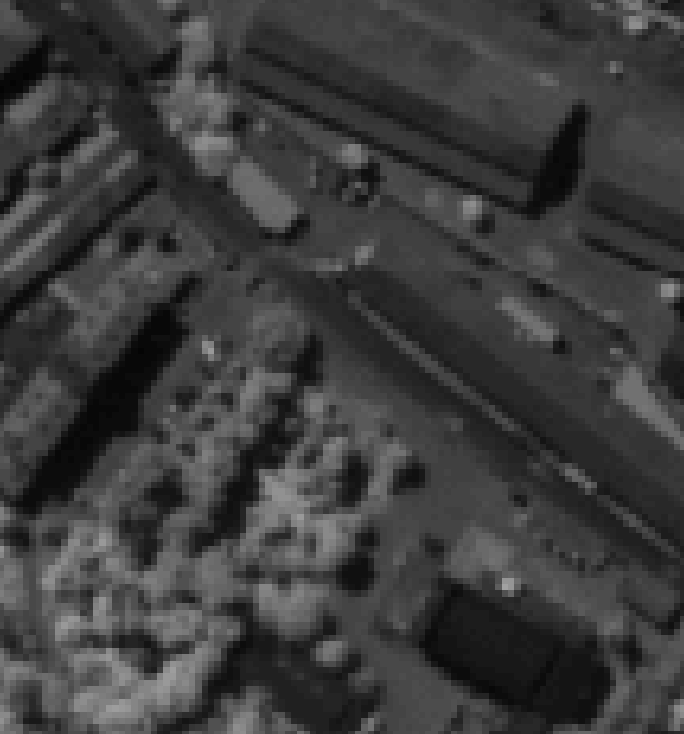}}
\subfigure[MAP1]{\includegraphics[width=\wi\linewidth]{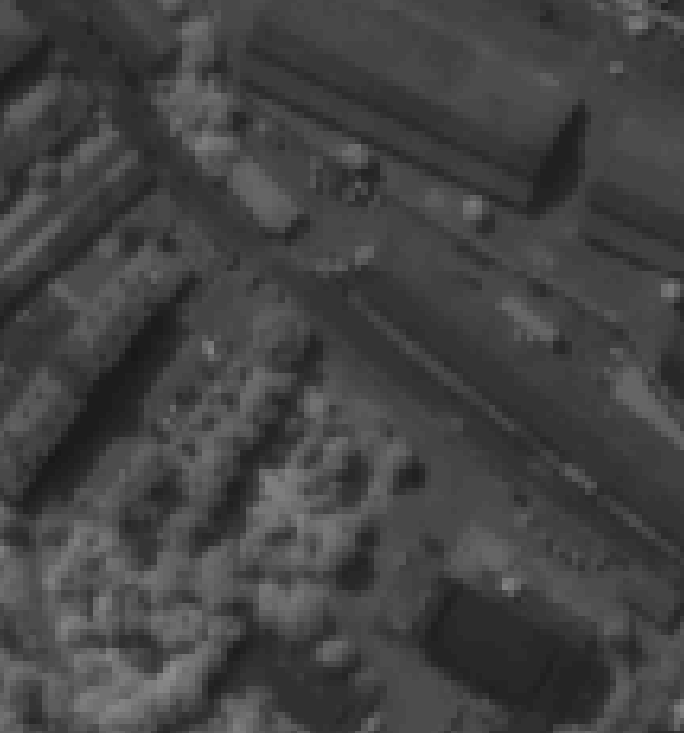}}
\subfigure[MAP2]{\includegraphics[width=\wi\linewidth]{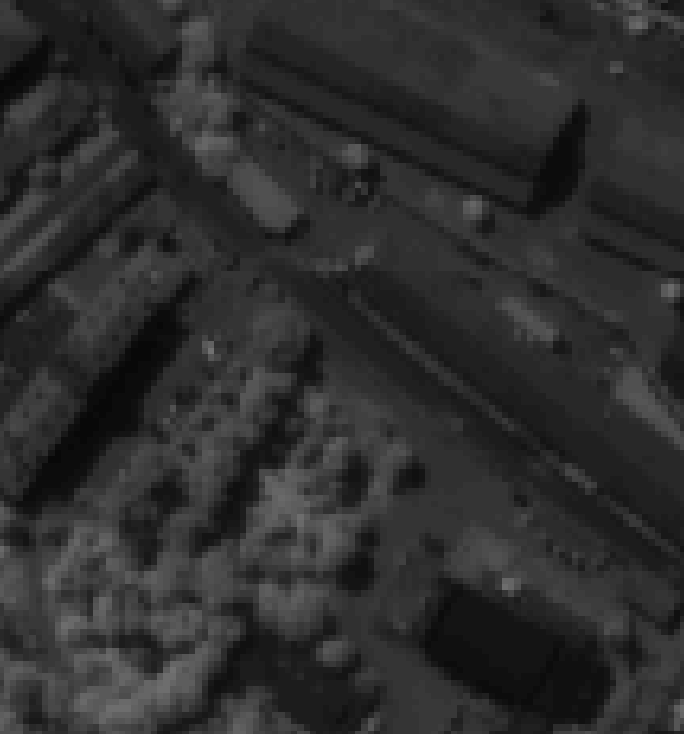}}
\subfigure[3D-CNN]{\includegraphics[width=\wi\linewidth]{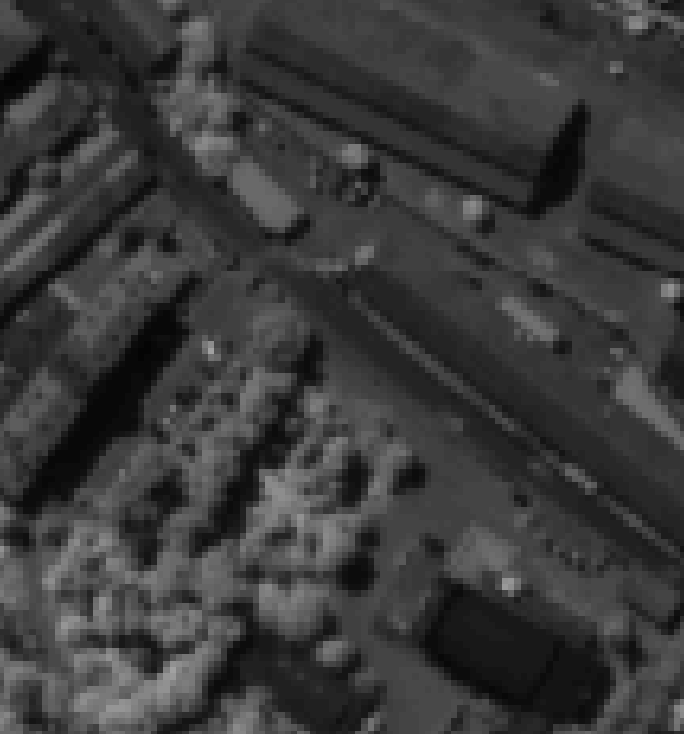}}
\caption{A subset of band 102 of the HS image is shown. (a) shows the interpolated HS image band, (b) is the reference band, (c) shows the image obtained using the MAP1\cite{Yifan} method, (d) shows the image obtained using the MAP2\cite{pwmbf} method and (e) shows the image obtained using the proposed method.}
\label{fig:band80}
\end{center}
\end{figure*}

\section{Conclusions}
In this paper, we proposed a new method for the fusion of MS and HS images using a 3D-CNN. An important component of the method is dimensionality reduction via PCA prior to the fusion. This decreases the computational cost significantly while having no impact on the quality of the fused image. In the presence of noise, the dimensionality reduction can improve the result. The proposed method is compared to two methods based on MAP estimation. Experiments using a simulated dataset demonstrated that the proposed method gives good results and is also tolerant to noise in the HS image. 

\ifCLASSOPTIONcaptionsoff
  \newpage
\fi

\bibliographystyle{IEEEtran}

{\small
\bibliography{refs}
}

\end{document}